\documentclass[lettersize,journal]{IEEEtran}
\usepackage{amsmath,amsfonts}
\usepackage{algorithmic}
\usepackage{algorithm}
\usepackage{array}
\usepackage[caption=false,font=footnotesize,labelfont=rm,textfont=rm]{subfig}
\usepackage{textcomp}
\usepackage{stfloats}
\usepackage{url}
\usepackage{verbatim}
\usepackage{graphicx}
\usepackage{cite}
\usepackage{booktabs}
\usepackage{xcolor}
\usepackage{makecell}
\usepackage{bbding}

\usepackage[T1]{fontenc} 

\makeatletter
\let\NAT@parse\undefined
\makeatother
\usepackage{hyperref}  
\newcommand{\fakepar}[1]{\smallbreak{}}

\begin{document}
\begin{small}
\title{Exploring Textual Semantics Diversity for Image Transmission in Semantic Communication Systems using Visual Language Model}
\end{small}
\author{Peishan Huang, Dong Li,~\IEEEmembership{Senior Member,~IEEE} 
\vspace{-1cm}
\thanks{The authors are with the School of Computer Science and Engineering, Macau University of Science and Technology, Macau, China. (e-mails: 3220006969@student.must.edu.mo;
dli@must.edu.mo).\emph{(Corresponding author: Dong Li.)} }
}

\maketitle

\begin{abstract}
In recent years, the rapid development of machine learning has brought reforms and challenges to traditional communication systems. Semantic communication has appeared as an effective strategy to effectively extract relevant semantic signals semantic segmentation labels and image features for image transmission. However, the insufficient number of extracted semantic features of images and inaccurate semantic descriptions will potentially result in a low reconstruction accuracy, which hinders the practical applications and still remains challenging for solving. In order to fill this gap, this paper proposes a multi-text transmission semantic communication (Multi-SC) system, which uses a feature fusion method and the visual language model (VLM) to assist in the transmission of image semantic signals. Unlike previous image transmission semantic communication systems, the proposed system divides the image into multiple blocks according to the object detection with the feature fusion and extracts multiple text information from the image using a modified large language and visual assistant (LLaVA), and combines semantic segmentation tags with semantic text for image recovery. The simulation results show that the proposed text semantics diversity scheme can significantly improve the reconstruction accuracy compared with related works.
\end{abstract}


\section{Introduction}
\label{intro}
\IEEEPARstart{W}{ith} the rapid development of communication technology at an ever increasing data rate, traditional syntactic information compression and propagation have approached the limitations of Shannon Capacity~\cite{7886994}. However, the emergence of semantic communication has brought the communication system beyond the boundaries of traditional communication, which focuses more on the restoration of semantic information from the transmitter to the receiver. For instance, when propagating high-quality images, traditional methods will require a large amount of resource consumption to convey the entire image. However, using natural language processing and computer vision algorithms to extract semantic features for image transmission will reduce the resource consumption to a certain extent and realize the image transmission process.

For semantic communication with image transmission, there have been a large number of works that have been paid attention to the joint source and channel encoding design, and there is an increasing interest in the generative image recovery. In particular, the language model has received a lot of attention, which can extract the textual semantic information of images and thus reduce the amount of information for transmission. Current research on textual semantic information for image recovery can be divided into two categories, one is utilized for image recovery~\cite{chen2024semanticcommunicationbasedlarge,10827002,10910032}, and the other one supervises the image recovery~\cite{liang2024imagegenerationmultimodulesemantic,10343094}. In~\cite{chen2024semanticcommunicationbasedlarge}, large language models were introduced into semantic communication systems to extract textual information and selectively transmit portrait segments of images. Although this approach can reduce the transmission of information to some extent, the system still faces a significant data burden when dealing with multiple portraits or large objects. In~\cite{10827002}, a new method was introduced to extract text information from images using a text-image model. The image was then reconstructed using this information, and a transformer-based text encoding and decoding model was designed to improve the robustness of text transmission over a noisy channel. In~\cite{10910032}, a new cross-modal semantic communication system based on a visual language model was proposed. This system utilized the Bootstrapping Language-Image Pretraining (BLIP) model to extract text semantics at the transmitter and reconstructed the image using the SD model at the receiver end. Although the text information can effectively reduce the amount of transmitted data, relying solely on the plain text will reduce the accuracy of reconstructed images. On the other hand, in~\cite{liang2024imagegenerationmultimodulesemantic}, a new method was proposed to extract both textual and image semantics at the transmitter, and the image semantics were used for image recovery and the textual semantics supervised and selected the most accurate generative image based on the generation and selection mechanism proposed in~\cite{10343094}.

In this paper, we focus on properly splitting the images and using the textual semantic information to recover the image. It should be noted that the reasonable division of images has a great impact on text feature extraction, and random division of images will reduce the amount of effective text extraction to a certain extent. In addition, the reconstruction accuracy of images is affected by the inherent randomness of the image generation model, and reducing the randomness is the focus of the previous works, which, however, is constrained by the limited textual semantic information. Thus, a nature question arises on how to explore more textual semantics for more accurate image recovery. In this regard, we propose a new scheme by exploring textual semantics diversity from multiple segmented parts of the image. Specifically, the contributions of this paper is summarized as
\begin{itemize}
    \item  We propose a multi-text transmission semantic communication (Multi-SC) system, where we adopt the Large Language and Vision Assistant (LLaVA) model~\cite{liu2023visual} to extract the text information from the image and transmit the text and the image embeddings to complete the image transmission process.
    \item We introduce a semantic segmentation module at the transmitter, which segments the main body of the image and performs embedding encoding. The receiver includes a text processing module to purify the text embedding contaminated by the channel noise (e.g., correcting spelling errors), thereby reconstructing the image more accurately.
    \item We introduce the feature fusion into the image block module at the transmitter, which divides the image into multiple blocks using the self-attention mechanism, and fuses similar blocks together through the feature fusion to improve the accuracy of text feature extraction and reduce the redundant useless text information.
    \item We evaluate the system performance using Bilingual Evaluation Understudy (BLEU)~\cite{papineni2002bleu}, ‌Learned Perceptual Image Patch Similarity (LPIPS)~\cite{8578166}, and cosine similarity, with the datasets VOC2012~\cite{pascal-voc-2012} COCO~\cite{DBLP:journals/corr/LinMBHPRDZ14} and Kodak24 from Kaggle. The results demonstrate significant improvements in image transmission compared to existing semantic communication systems.
\end{itemize}

The rest of the paper is organized as follows. Section~\ref{sys_model} introduces the system framework, Section~\ref{alg} describes the relevant models, Section~\ref{sim} outlines the experimental setup and results, and Section~\ref{conclusion} summarizes the paper.

\section{system model}
\label{sys_model}

\begin{figure*}[htbp]\vspace{-1cm}
	\centering
	\includegraphics[scale=0.18]{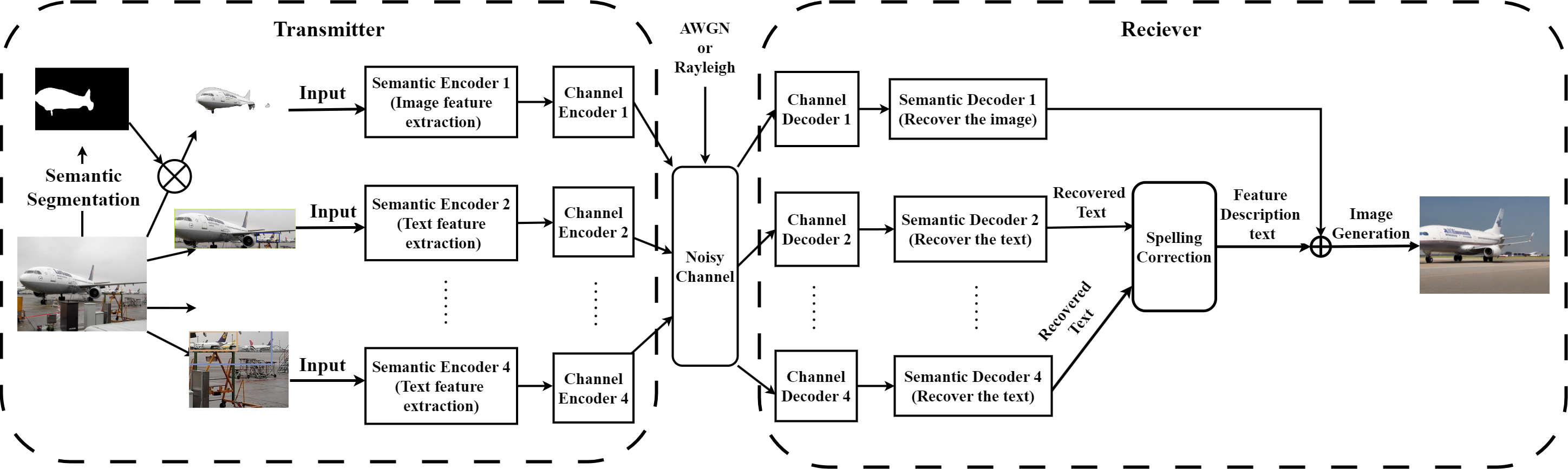}
	\caption{The framework of multi-text enabled image transmission semantic communication system.}
	\label{vlm}
\end{figure*}

In this paper, we consider an image transmission semantic communication system assisted by visual language model, and the system framework is shown in Fig.\ref{vlm}, where the transmitter mainly realizes the process of extracting text and image features, and the receiver realizes the process of recovering the features affected by the noise and reconstructing the image by using the features. The specific steps are described in details as follows. 

\subsection{Transmitter}
At the transmitter, the first step is to cut out the main part of the image to be transmitted through the Fully Convolutional Networks (FCN)~\cite{long2015fully} $ N_{FCN}$, and the expression of the process is
\begin{equation}
    i_m = N_{FCN}(i),
    \label{fcn}
\end{equation}
where $i$ is the image to be transmitted, and $i_m$ is a slice of the main part of the image, and the slice will be transmitted. Meanwhile, the image to be transmitted will be divided into multiple sub-images $i_n,n=\{1,2,3,\dots\}$ using an object detection model. The process is expressed as
\begin{equation}
    i_n = N_{detection}(i),
    \label{yolo}
\end{equation}

In this paper, the divided images will be delivered to different devices and the LLaVA~\cite{liu2023visual} $F_{I2T}$ to extract text features for the second step. The process is shown as 
\begin{equation}
    e_n=F_{I2T}(q,i_n),
    \label{LLaVA_enceode}
\end{equation}
where $q$ is a fixed query, and $e_n$ is the text embedding of the description text. The device will then transmit the parameters associated with the text feature to the receiver via the orthogonal noisy channels. The feature extraction of the text information and a small amount of image information can reduce a large amount of redundant data transmission and the possibility of misrepresentation of the subject information.

\subsection{Receiver}
For simplicity of analysis, we consider the additive white Gaussian noise (AWGN) channel and Rayleigh channel, and the received signal $\hat{z_i}$ is given by
\begin{equation}
    \hat{z_i}=h_iz_i+n_i,
    \label{snr}
\end{equation}

\noindent where $z_i$ is the transmitted signal, $h_i$ is the channel gain and $n_i$ is the noise distribution of the user $i$. When the signal was transmitted under AWGN channel, the channel gain $h$ equals to 1 and $n$ follows the Gaussian distribution $(0,\sigma^2I)$. As for Rayleigh channel, $h$ is a complex channel gain, which follows the complex Gaussian distribution, and the noise distribution $n$ is similar to that of the AWGN channel.

At the receiver, the receiver will decode the received information using an image decoder and a text decoder, respectively, and the decoding process is given by
\begin{equation}
    t_n'=F_{I2T}(e_n'),
    \label{LLaVA decoder}
\end{equation}

\noindent where $t_n'$ is the noisy text and $e_n'$ is the noisy text embedding.

The receiver will fuse all the received text to form a complete the text feature. For the text features, the channel noise will cause the text to be misspelled differently compared with the source text. To this end, Bidirectional and Auto-Regressive Transformers (BART)~\cite{lewis2019bart}$N_{BART}$ is used to correct the spelling of the text. The spelling correction process is given by
\begin{equation}
    t_n=N_{BART}(t_n'),
    \label{BART}
\end{equation}

\noindent where $t_n$ is the description text of the image $i_n$ with spelling correction.

The processed text will be input into Image Prompt Adapter (IP-Adapter)~\cite{ye2023ip-adapter}$F_{T2I}$ together with the decoded image for image generation to complete the reconstruction of the complete image, and the reconstruction process given by
\begin{equation}
    i_{recon}=F_{T2I}(t_1,t_2,\dots,t_n,i_m'),
    \label{IPAdapter}
\end{equation}
\noindent where $i_{recon}$ is the reconstruction image, and $i_m'$ is the slice of the main part of the image with noise. The whole process is summarized in Algorithm~\ref{alg:alg1}.

\begin{algorithm}[H]
	\caption{Proposed Multi-Text Semantic Communication with Feature Fusion Split (Multi-SC with Feature Fusion Split)}\label{alg:alg1}
	\begin{algorithmic}
		\STATE 
		\STATE {\textbf{Require}:} 
            \STATE \hspace{0.5cm} An image $i$ from dataset $I$, pretrained LLaVA model $F_{I2T}$, pretrained IP-Adapter $F_{T2I}$, BART $N_{BART}$ with pretrained parameters $\theta_{B}$, FCN $N_{FCN}$ with initialized parameters $\theta_{FCN}$ and the threshold $\epsilon$.
            \STATE 
		\STATE {\textbf{Transmitter}:}
		\STATE \hspace{0.5cm}1. Compute a main object $i_{m}$ with $N_{FCN}$ and segregate $i_{m}$ from $i$ with~(\ref{fcn}).
		\STATE \hspace{0.5cm}2. Divide $i$ into several parts $i_{a}$, $i_{b}$, $i_{c}$, $i_{d}$, $i_{e}$ using~(\ref{yolo}). Compute the distance between each sub-images, e.g. $d_{ab}$, $d_{bc}$. If $d_{ab}<\epsilon$, $i_{a}$, $i_{b}$ are fused into picture $i_{1}$, otherwise $i_{a}$ relabels as $i_{1}$ and $i_{b}$ relabels as $i_{2}$.
            \STATE \hspace{0.5cm}3. Extract text embedding $e_{1}$, $e_{2}$, $e_{3}$ with $F_{I2T}$ using~(\ref{LLaVA_enceode}) from $i_{1}$, $i_{2}$, $i_{3}$.
		\STATE \hspace{0.5cm}4. Transmit $i_{m}$, $e_{1}$, $e_{2}$, $e_{3}$ through physical channels with AWGN and Rayleigh, respectively.
		\STATE 
		\STATE {\textbf{Receiver}:}
		\STATE \hspace{0.5cm}1. Receive $i_{m}'$, $e_{1}'$, $e_{2}'$, $e_{3}'$.
		\STATE \hspace{0.5cm}2. Recover text $t_{1}'$, $t_{2}'$, $t_{3}'$ with $F_{I2T}$ according to (\ref{LLaVA decoder}) and use $N_{BART}$ to correct the spelling $t_{1}$, $t_{2}$, $t_{3}$ with (\ref{BART}).
		\STATE \hspace{0.5cm}3. Reconstruct image $i_{recon}$ using $F_{T2I}$ according to (\ref{IPAdapter}) with $i_{m}'$, $t_{1}$, $t_{2}$, $t_{3}$.
	\end{algorithmic}
	\label{multi-sc}
\end{algorithm}

\section{Transceiver Architecture Design}
\label{alg}
In this section, we will present the details of transceiver architecture design as shown in the previous section.  
\subsection{Semantic segmentation}
In image transmission, the content extraction of the main object is essential, and the loss of the main content will make the receiver unable to accurately restore the original image information. Therefore, the FCN is introduced to crop the main object in the image, which is a deep learning architecture for image recognition and segmentation. The main contribution of FCN is to apply CNN to full-image segmentation tasks and to be able to output segmentation maps with the same resolution as the input image. The structure of the FCN is to replace the fully connected layer in the traditional Convolutional Neural Networks (CNN) with a convolutional layer, which allows the model to process images of any size and maintain the spatial dimension of the output. The FCN increases the size of the feature map by upsampling so that the output has the same resolution as the input image. In this paper, the FCN model will be used for image segmentation, that is, the main object in the transmitted image will be identified and segmented as a part of the transmitted information. The Negative Log-Likelihood Loss (NLL) loss function is as follows:

\begin{equation}
    NLL(p(y|x),y)=-\log p(y|x),
    \label{fcn_loss}
\end{equation}

\noindent where $x$ is an input sequence, $y$ is a true label and $p(y|x)$ is the predicted label distribution of the model. 

\subsection{Feature fusion}
In the transmitter, we perform feature fusion operations on all detected segmented sub-images. Since too many sub-images with duplicate features will lead to redundant information and reduce the information efficiency. Therefore, we adopt the feature fusion method to improve the efficiency of information transmission. For sub-images that repeatedly appear the same object, feature fusion is necessary to use on these sub-images from the original image. For example, $i_{a}$ and $i_{b}$ are sub-images of repeating information, and the Euclidean distance of their center points is calculated by:
\begin{equation}
    d_{ab}=\sqrt{(x_a-x_b)^2+(y_a-y_b)^2},
    \label{dis}
\end{equation}
\noindent when the gap $d_{ab}$ is less than the threshold $\epsilon$, the sub-images $i_{a}$, $i_{b}$ are fused into a new sub-image $i_{1}$. Otherwise, when the gap $d_{ab}$ is greater than the threshold $\epsilon$, the two target images $i_{a}$, $i_{b}$ will be divided into two separate sub-images $i_{1}, i_{2}$.

\subsection{Text extraction}
\begin{figure}[htpb]
	\centering
	\includegraphics[scale=0.15]{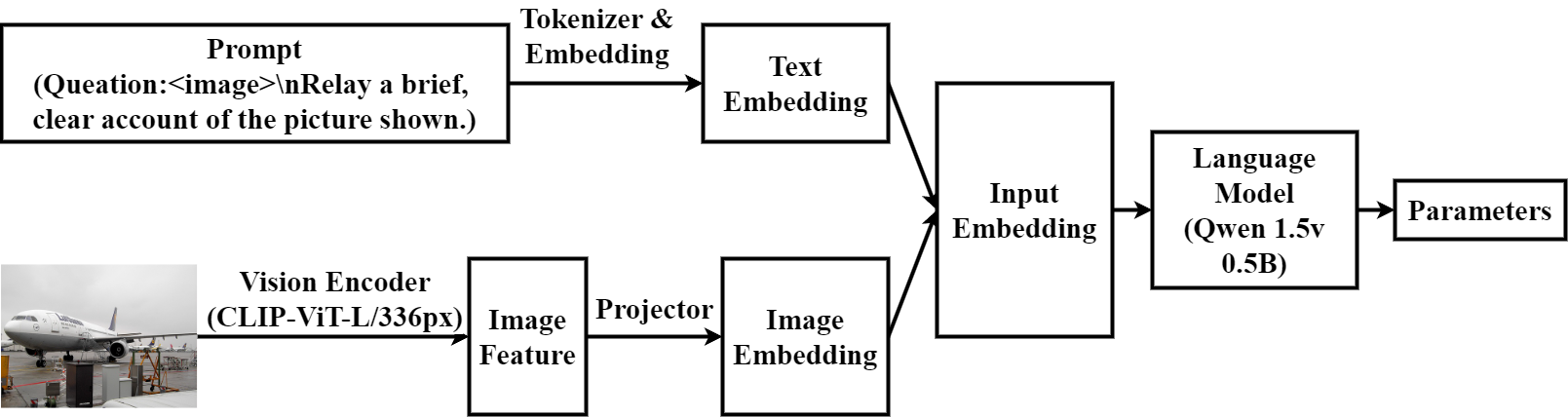}
	\caption{The framework of semantic encoder with LLaVA.}
	\label{llava}
\end{figure}

After the sub-image partitioning is completed, the accurate extraction of text information is the next key semantic encoding step. In the semantic encoder, we consider the LLaVA model for text extraction as shown in Fig.~\ref{llava}. LLaVA is a large multimodal visual language model trained end-to-end, which connects visual encoding and large language models to achieve general visual and language understanding. In Fig.~\ref{llava}, the LLaVA model uses a vision encoder to get the features $Z_v=g(X_v)$ from images. Then, we apply a trainable projection matrix W to convert features $Z_v$ into language embedding tokens $H_v$ as $H_v=W\cdot Z_v$. The language model embeds the text. The text processed by the language model is "<image>$\backslash$nRelay a brief, clear account of the picture shown." Then, LLaVA uses a simple linear layer to align the two embeddings, that is, the image embedding will be placed in the embedding position corresponding to "<image>" in the text.

\subsection{Spelling correction}
Due to the presence of noise in the channel, distorted text information is inevitable. therefore, in the reciever, BART is introduced as a spelling correction model, which is a pre-trained model that combines the characteristics of BART~\cite{lewis2019bart}. BART uses an encoder-decoder structure based on the standard Transformer architecture, in which the encoder inherits the bidirectional encoder of Bidirectional Encoder Representations from Transformers (BERT) and the decoder inherits the autoregressive decoder of Generative Pre-trained Transformer (GPT), which enables the algorithm to understand the text context and the generated coherent text. BART is trained by destroying the text (such as adding noise, masking, etc.) and then learning to reconstruct the original text. Meanwhile, BART is trained using a variety of different noise methods to reduce the model's dependence on the structured information, enhance the model's understanding of text semantics, and be more conducive to the restoring damaged text. The Gaussian Error Linear Units (GeLU) activation function is used in the BART model instead of the default Rectified Linear Unit (ReLU) in the Transformer.

\subsection{Image Renconstruction}
Finally, the accuracy of image reconstruction is a key part of image transmission. IP-Adapter, is an adapter specifically designed for pre-trained text-to-image diffusion models as an image reconostruction model in this paper, which allows the model to generate images using image prompts. IP-Adapter can receive text prompts and image prompts simultaneously to generate corresponding images, and the model has a higher degree of freedom and flexibility. The core design of the IP-Adapter includes an image encoder and a decoupled cross-attention mechanism, which enables it to embed image features into a pre-trained text-to-image diffusion model, thereby enabling the generation of image prompts. In IP-Adapter, text features are input into the pre-trained model through the cross-attention. The output of the text cross-attention is given by

\begin{equation}
    Z_{text}=Attention(Q_t,K_t,V_t)=Softmax\frac{Q_tK_t^T}{\sqrt{d_t}}V_t,
    \label{ip-text}
\end{equation}

\noindent where $Q_t=Z_tW_q$, $K_t=c_tW_k$, $V_t=c_tW_v$, $Z_t$ is a query of latent space noise feature map, $c_t$ is the text feature and $W_q$, $W_k$, $W_v$ is the projection matrix of query, key, value respectively. The output of the image cross-attention is given by

\begin{equation}
    Z_{img}=Attention(Q_i,K_i,V_i)=Softmax\frac{Q_iK_i^T}{\sqrt{d_i}}V_i,
    \label{ip-img}
\end{equation}

\noindent where $Q_i=Z_iW_q$, $K_i=c_iW_k$, $V_i=c_iW_v$, $Z_i$ is a query of latent space noise feature map, $c_i$ is the image feature and $W_q$, $W_k$, $W_v$ is the projection matrix of query, key, value respectively.

Finally, we combine the output of the image cross-attention with the output of the text cross-attention as $Z_{new}=Z_{text}+Z_{img}$. Since the cross attention of image and text is decoupled, the weight coefficient can be used to adjust the weight of each on the final output as follows

\begin{equation}
    Z_{new}=Z_{text}+\lambda Z_{img},
    \label{ip}
\end{equation}

\noindent where $\lambda$ is the weight. It should be noted that when $\lambda$ is equal to 0, IP-Adapter becomes a normal text-to-image model. The loss function of IP-Adapter model is given by

\begin{equation}
    L_{simple}=E_{x_0,\epsilon,c_t,c_i,t}\|\epsilon-\epsilon_{\theta}(x_t,c_t,c_i,t)\|^2,
    \label{ip-loss}
\end{equation}

\noindent where $x_0$ is the real data, $t$ is the step and $x_t=\alpha_tx_0+\sigma_t\epsilon$ is the noisy data of t-th steps. $\alpha_t$ and $\sigma_t$ are calculated in advance based on hyperparameters.

\section{Simulation}
\label{sim}
In this section, we will evaluate the proposed multi-text transmission semantic communication system. The simulation environment is Tesla V100-SXM2-32GB, and the model is built with 2.1.2 PyTorch and 3.10 Python on Ubuntu 22.04.

\subsection{Simulation Settings}

The datasets used in the simulation are VOC2012, COCO and Kodak24. Due to the limited computing resources in the simulation environment, we make the LLaVA model lightweight and replace the language model with a smaller model, Qwen1.5-0.5B. In this simulation, we select some samples for prediction comparison in Table~\ref{lightweight comparison}. The accuracy of lightweight is not much different from the original model, and the text generated by the lightweight model is reduced by 4 times, which is more suitable for our scenario. In addition, lightweight runtime is more appropriate for our current resource conditions, which saves 2 times the computing resources and speeds up by nearly 4 times on the training time.

\begin{table}[h]
	\centering
	\caption{Model lightweight comparison}
	\setlength{\tabcolsep}{3mm}{
		\begin{tabular}{cccc}
			\toprule 
                \textbf{Model} & \textbf{BLEU}& \textbf{Text length}& \textbf{training time}   \\
                 \midrule
                Qwen1.5-0.5B & 0.1485 & 42  & 5 hours (with 4 V100) \\
                Vicuna1.5-7B & 0.1451 & 190 & 20 hours (with 8 V100) \\
			\bottomrule
	\end{tabular}}
	\label{lightweight comparison}%
\end{table}%

Regarding the training of the BART model, we reconstruct a new training set based on the scenario in this paper. All the images in the image dataset are put into the pretrained LLaVA model to generate the corresponding text, since the text generated by Vicuna is longer, which is more conducive to the training of the BART model. However, due to the limited number of images and the small amount of text, we split the image into multiple pieces to generate the corresponding text. Finally, these text data are combined and randomly sampled to generate a training set.

\subsection{Performance Analysis}
In the simulation, we compare the proposed model Multi-SC with feature fusion split and the baseline Multi-SC with random split with several methods: BPG-LDPC, deep Joint Source-Channel Coding (deep JSCC)~\cite{8683463}, and Vision-Language Model-based Cross-modal Semantic Communication (VLM-CSC)~\cite{10910032} with stable diffusion. In addition, we include a single-text framework for comparison. BPG-LDPC is a method that combines the classic image compression using Better Portable Graphics (BPG)~\cite{9349289} with Low-Density Parity-Check (LDPC) codes~\cite{1057683}.

\begin{figure}[htpb]\vspace{-0.7cm}
	\centering
	\subfloat[Cosine Similarity versus SNR.]{\includegraphics[width=0.5\linewidth]{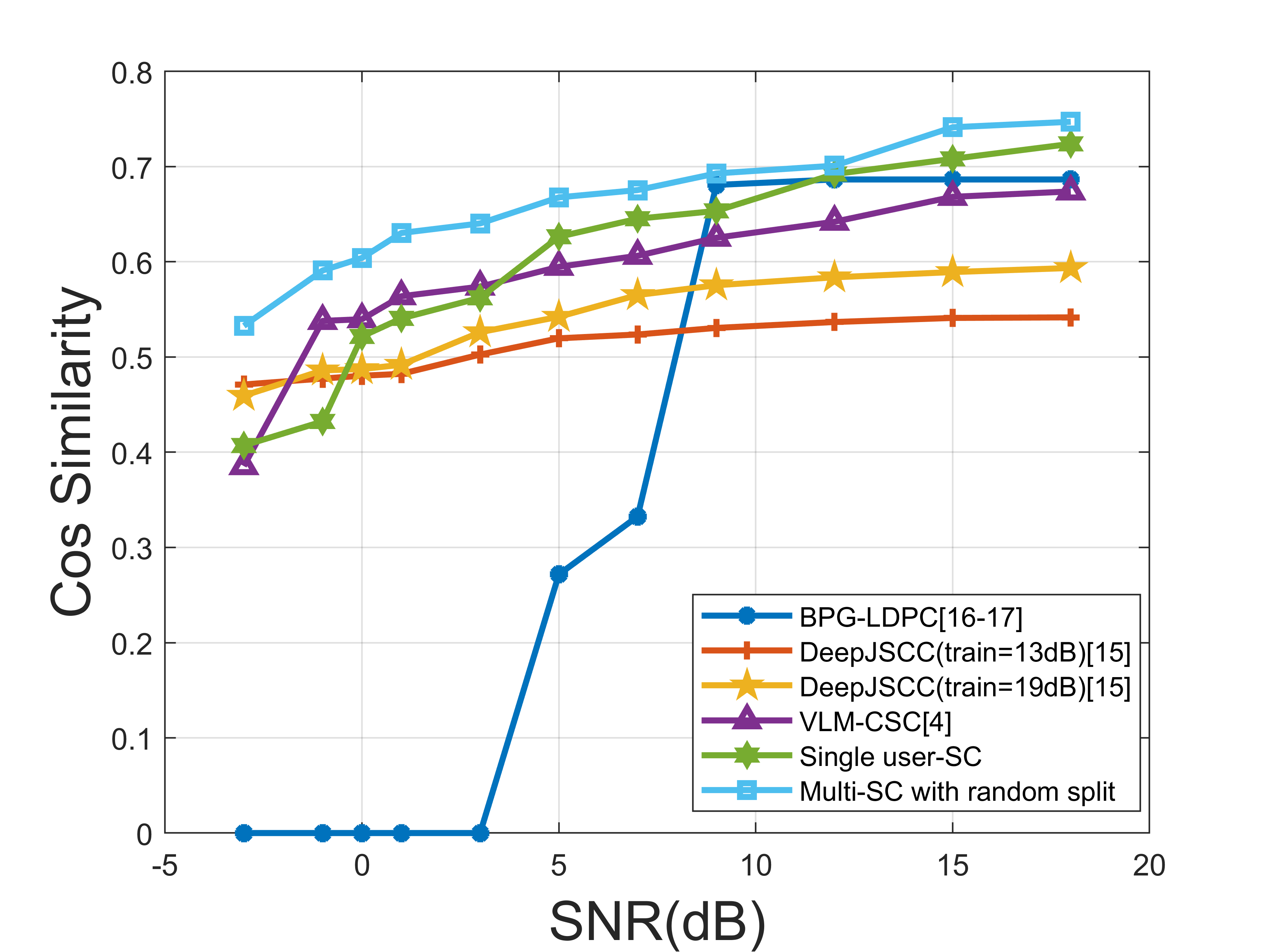}%
		\label{coss_kodak}}
	\hfil
	\subfloat[LPIPS versus SNR.]{\includegraphics[width=0.5\linewidth]{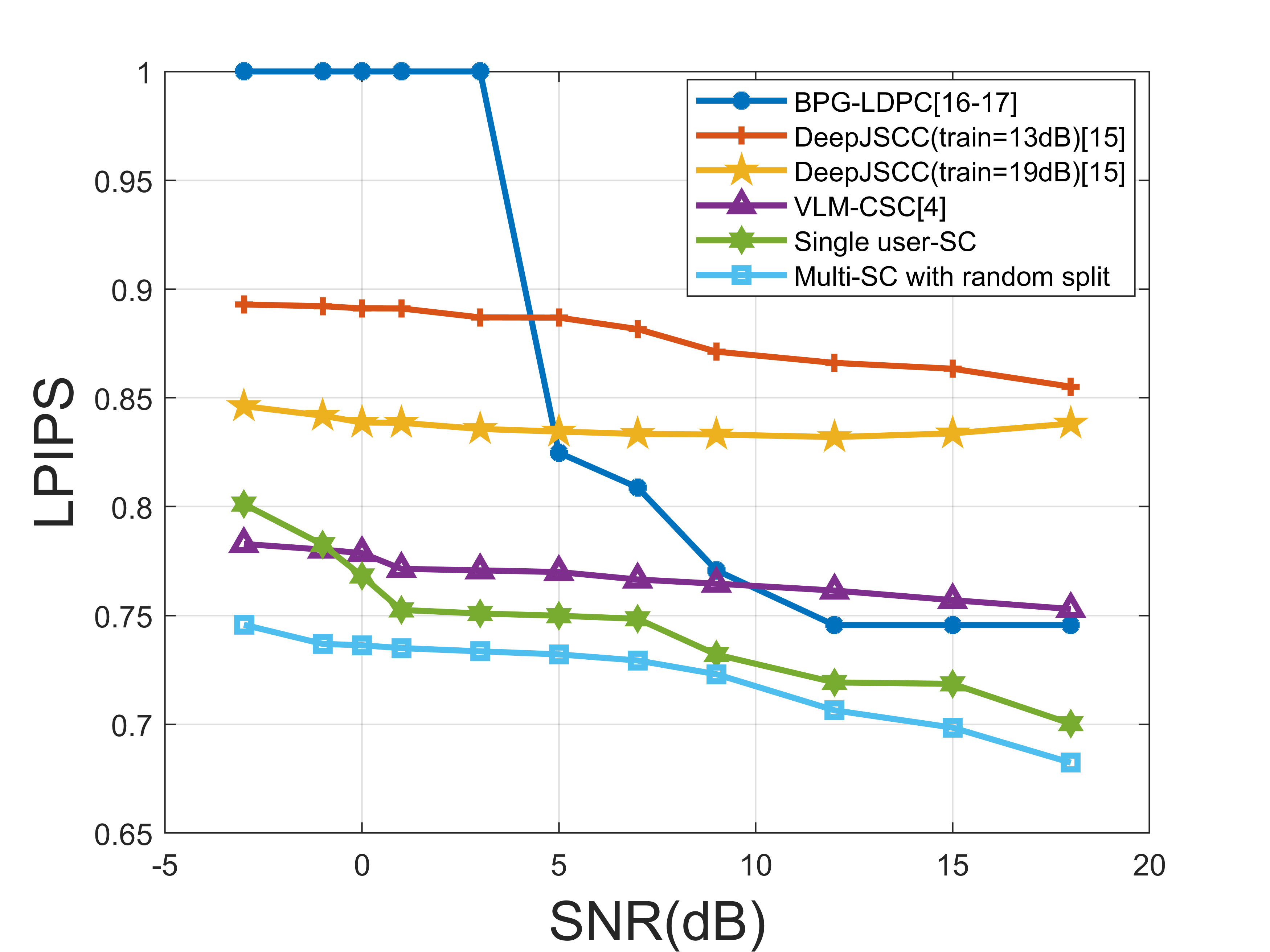}%
		\label{lpips_kodak}}
	
	\caption{The results of cosine similarity and LPIPS compared with BPG-LDPC, deep JSCC, VLM-CSC, single text-SC, Multi-SC with random split and baseline Multi-SC with feature fusion split on Kodak24 under AWGN.}
	\label{fig_kodak}
\end{figure}

\begin{figure}[htpb]\vspace{-0.7cm}
	\centering
	\subfloat[Cosine Similarity versus SNR.]{\includegraphics[width=0.5\linewidth]{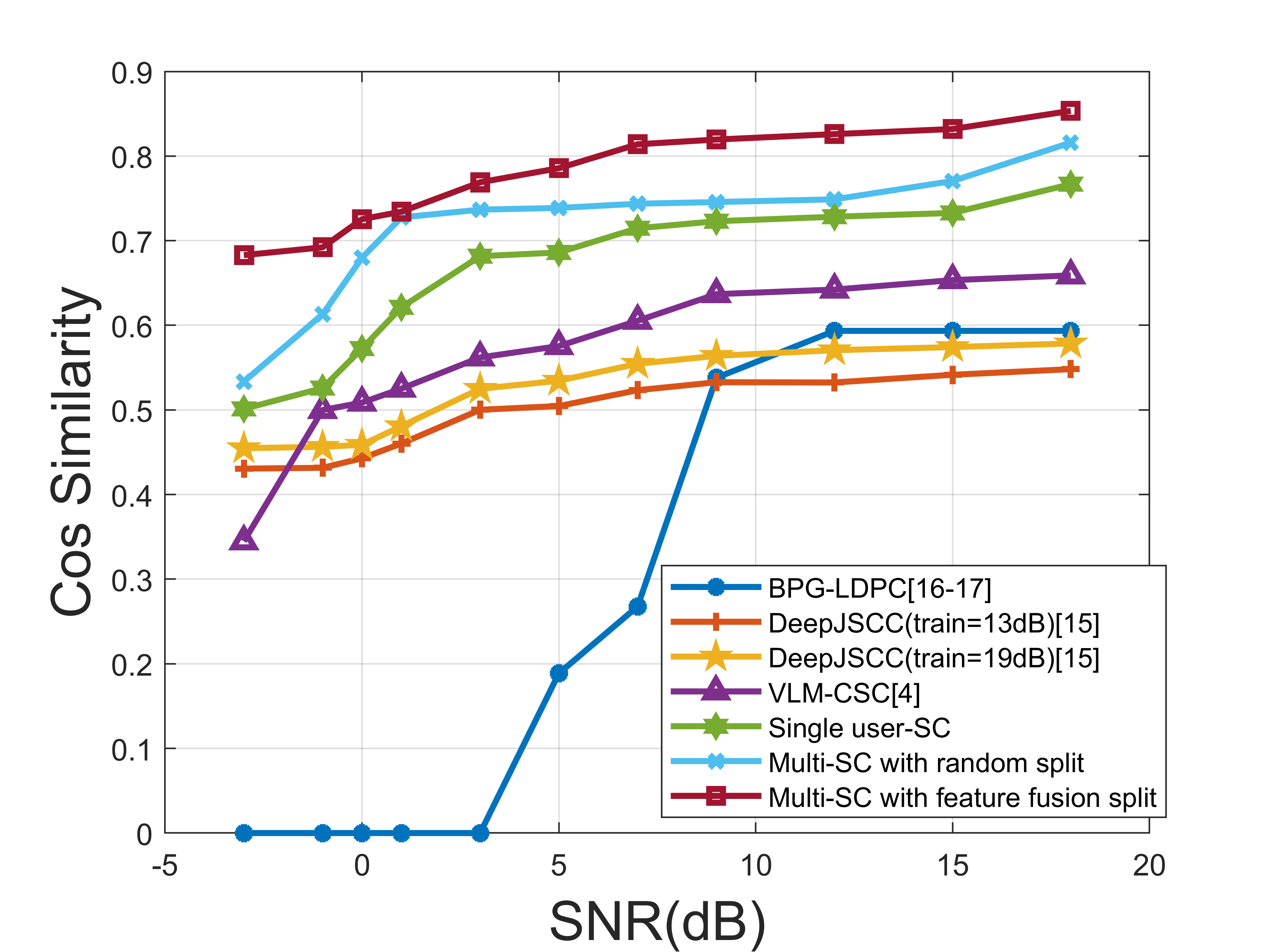}%
		\label{coss_voc}}
	\hfil
	\subfloat[LPIPS versus SNR.]{\includegraphics[width=0.5\linewidth]{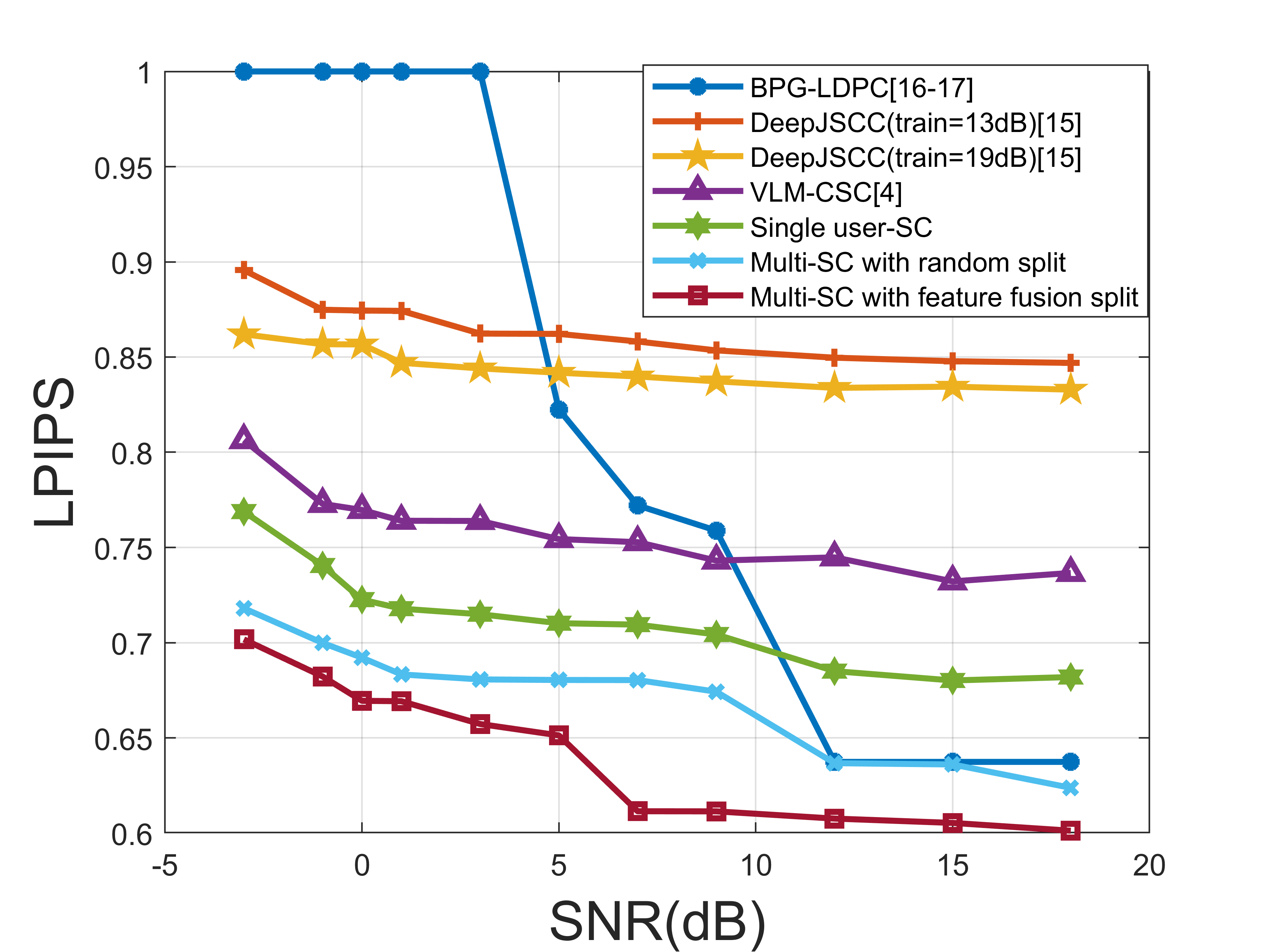}%
		\label{lpips_voc}}
	
	\caption{The results of cosine similarity and LPIPS compared with BPG-LDPC, deep JSCC, VLM-CSC, single text-SC, Multi-SC with random split and baseline Multi-SC with feature fusion split on VOC2012 under AWGN.}
	\label{fig_voc}
\end{figure}

\begin{figure}[htpb]\vspace{-0.7cm}
	\centering
	\subfloat[Cosine Similarity versus SNR.]{\includegraphics[width=0.5\linewidth]{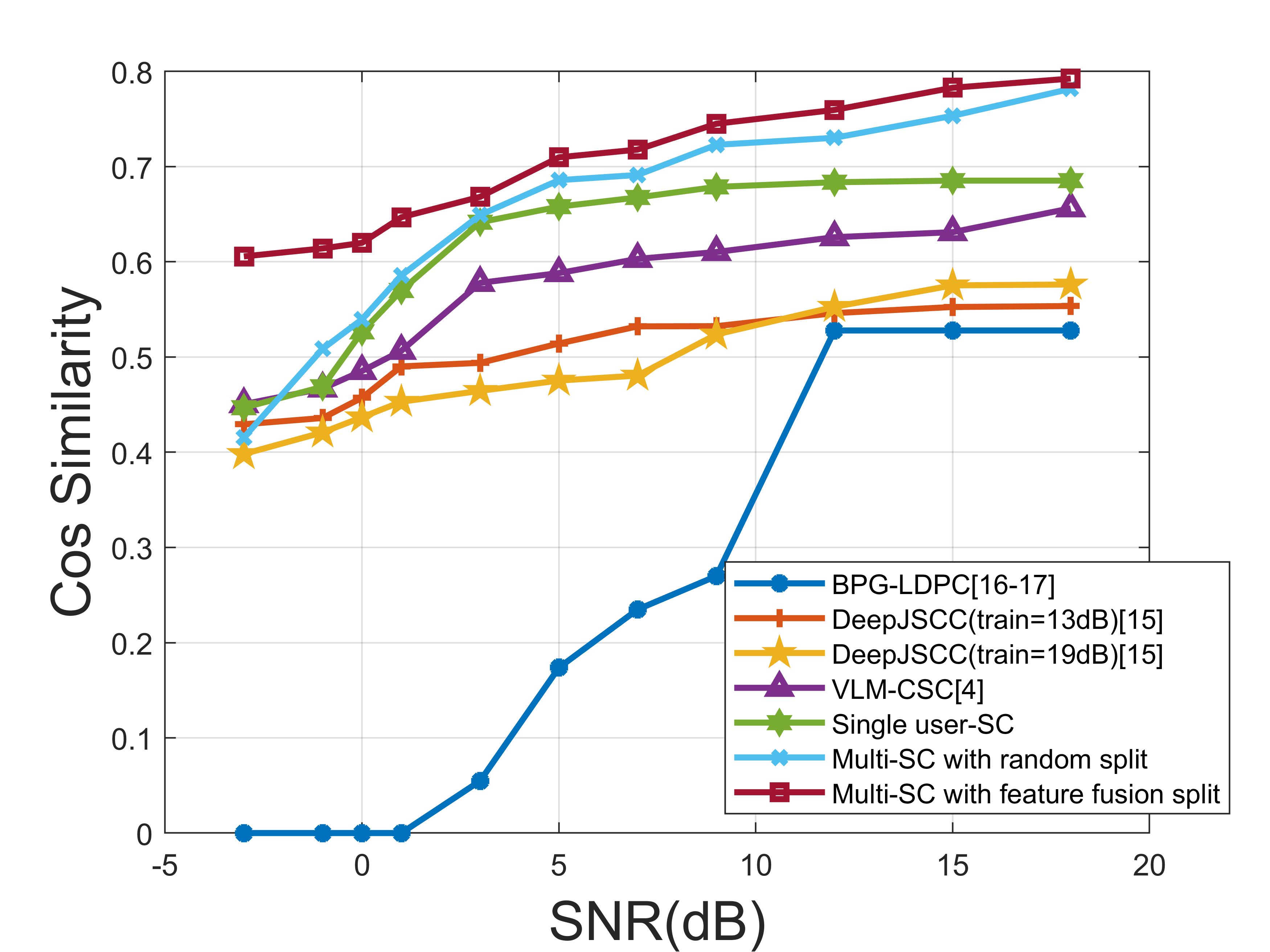}%
		\label{coss_coco}}
	\hfil
	\subfloat[LPIPS versus SNR.]{\includegraphics[width=0.5\linewidth]{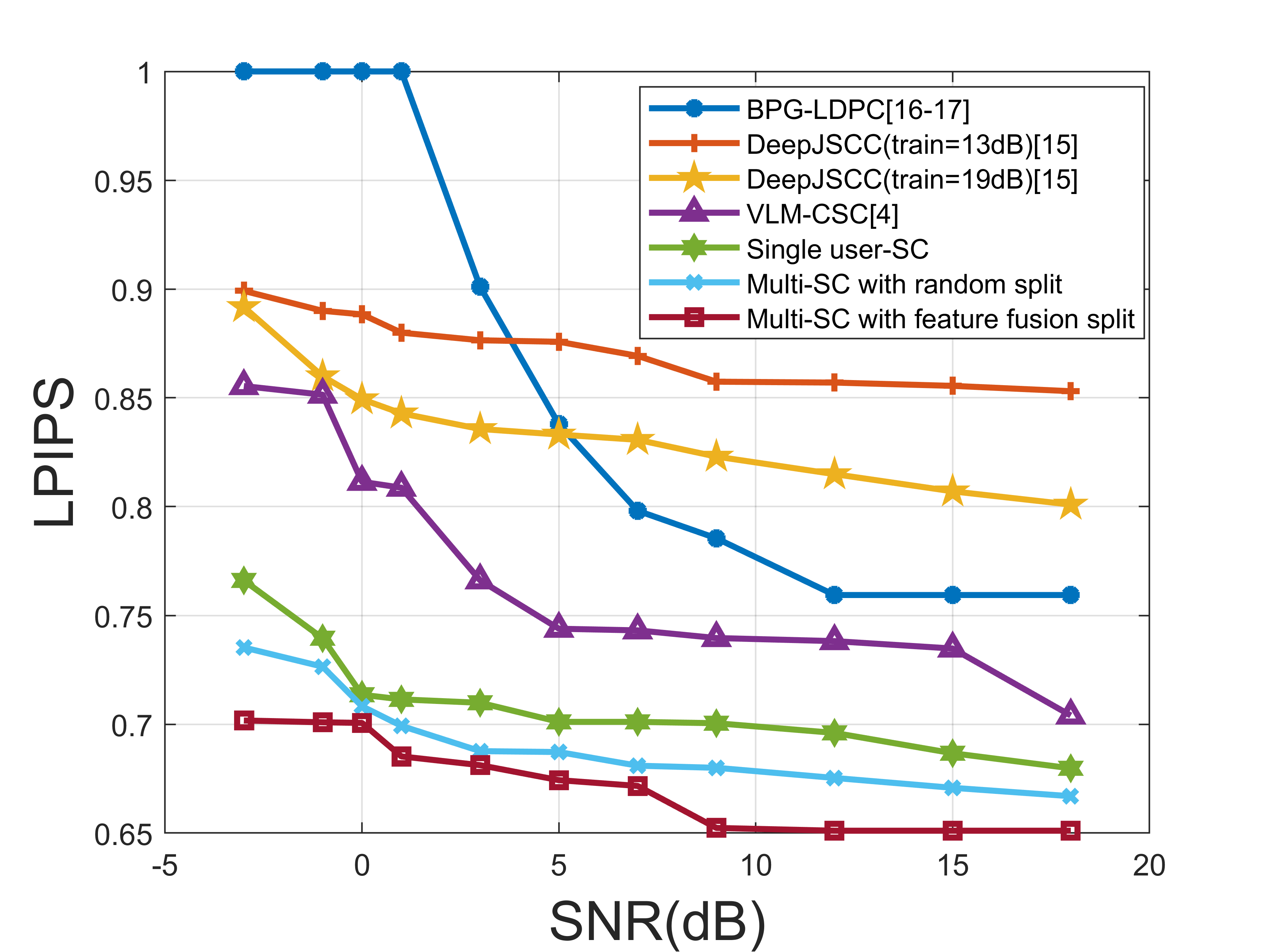}%
		\label{lpips_coco}}
	
	\caption{The results of cosine similarity and LPIPS compared with BPG-LDPC, deep JSCC, VLM-CSC, single text-SC, Multi-SC with random split and baseline Multi-SC with feature fusion split on COCO under AWGN.}
	\label{fig_coco}
\end{figure}

\begin{figure}[htpb]\vspace{-1cm}
	\centering
	\subfloat[Cosine Similarity versus SNR.]{\includegraphics[width=0.5\linewidth]{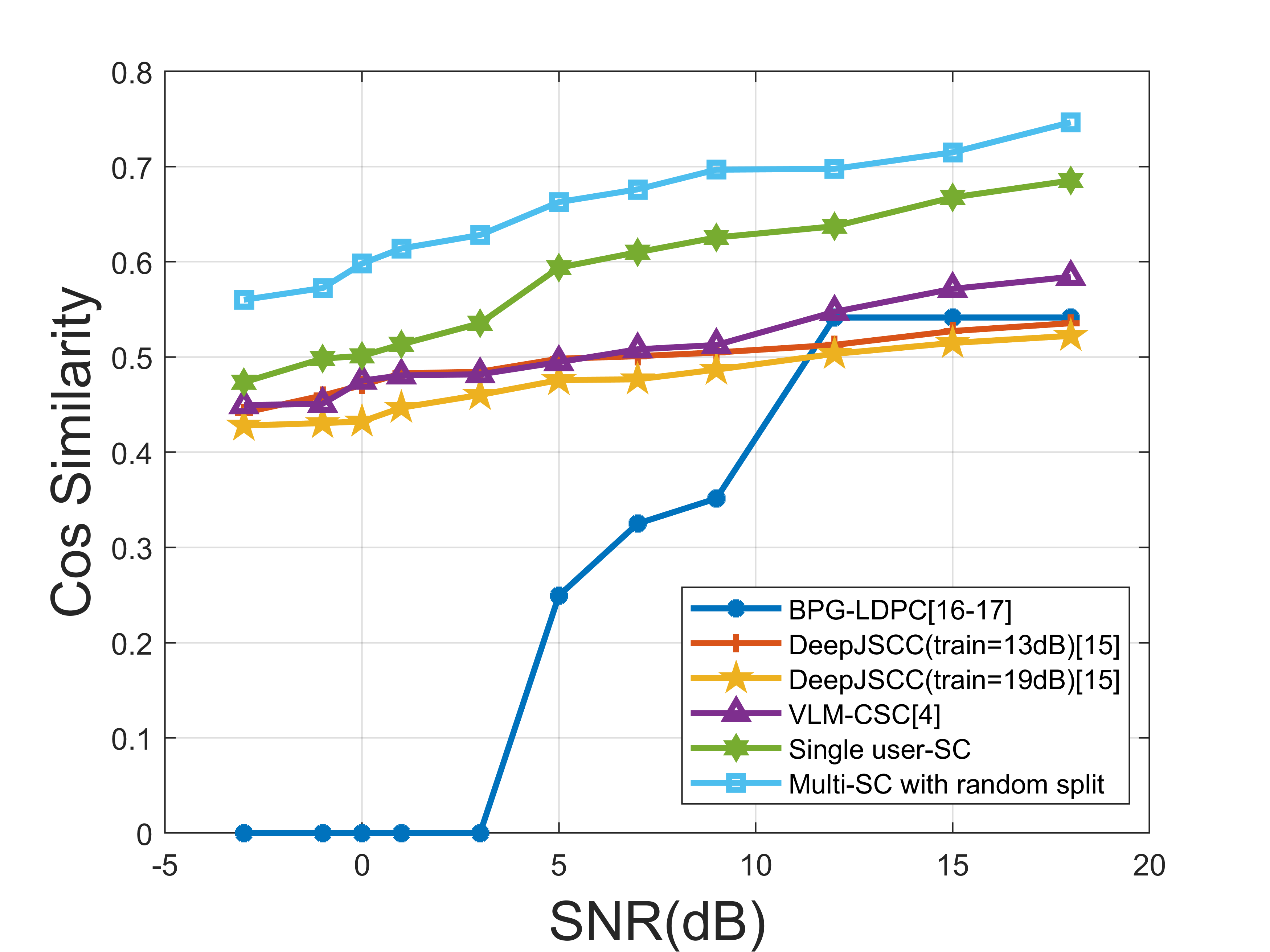}%
		\label{coss_kodak_ray}}
	\hfil
	\subfloat[LPIPS versus SNR.]{\includegraphics[width=0.5\linewidth]{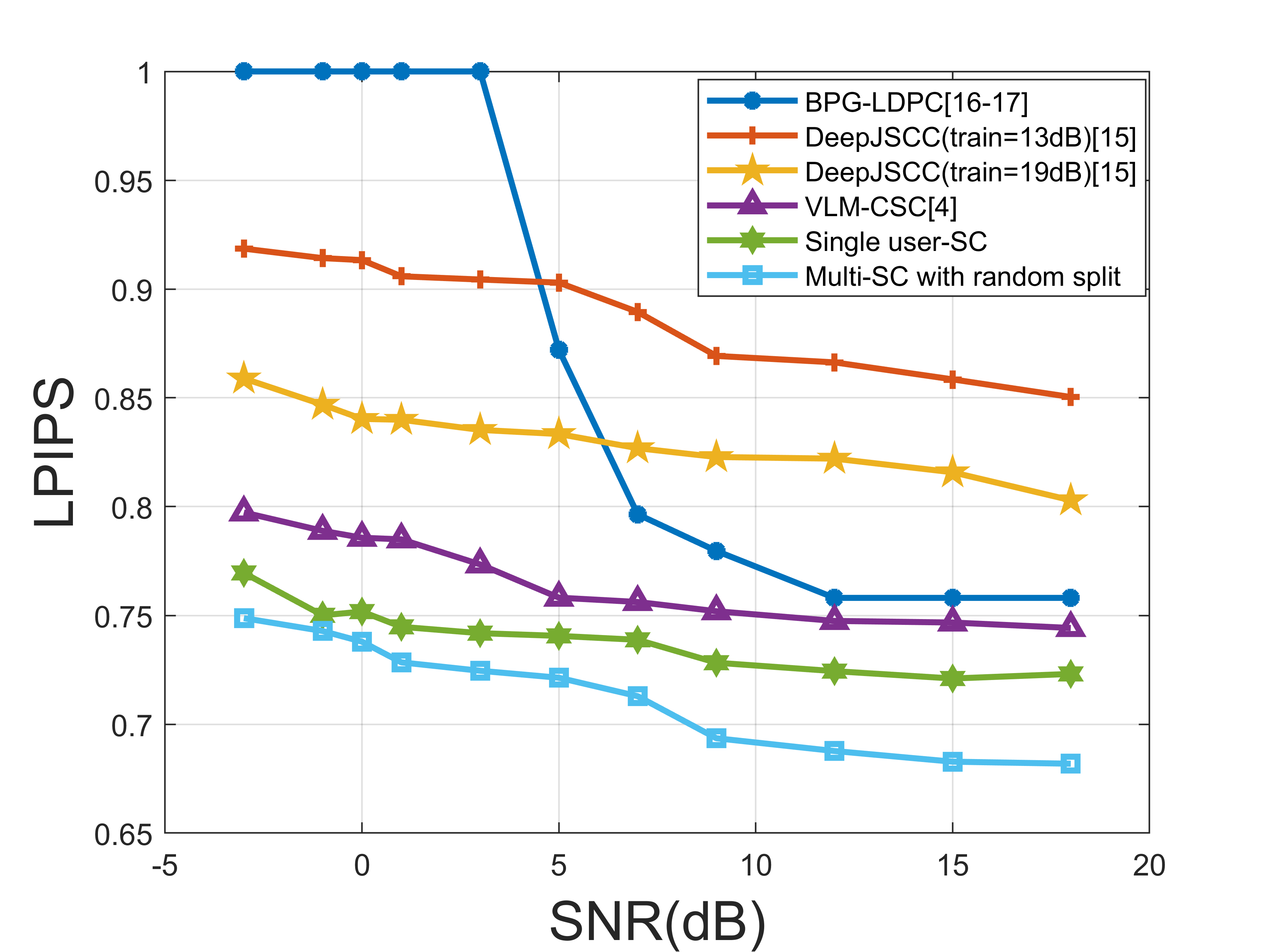}%
		\label{lpips_kodak_ray}}
	
	\caption{The results of cosine similarity and LPIPS compared with BPG-LDPC, deep JSCC, VLM-CSC, single text-SC, Multi-SC with random split and baseline Multi-SC with feature fusion split on Kodak24 under Rayleigh.}
	\label{fig_kodak_ray}
\end{figure}

\begin{figure}[htpb]\vspace{-0.7cm}
	\centering
	\subfloat[Cosine Similarity versus SNR.]{\includegraphics[width=0.5\linewidth]{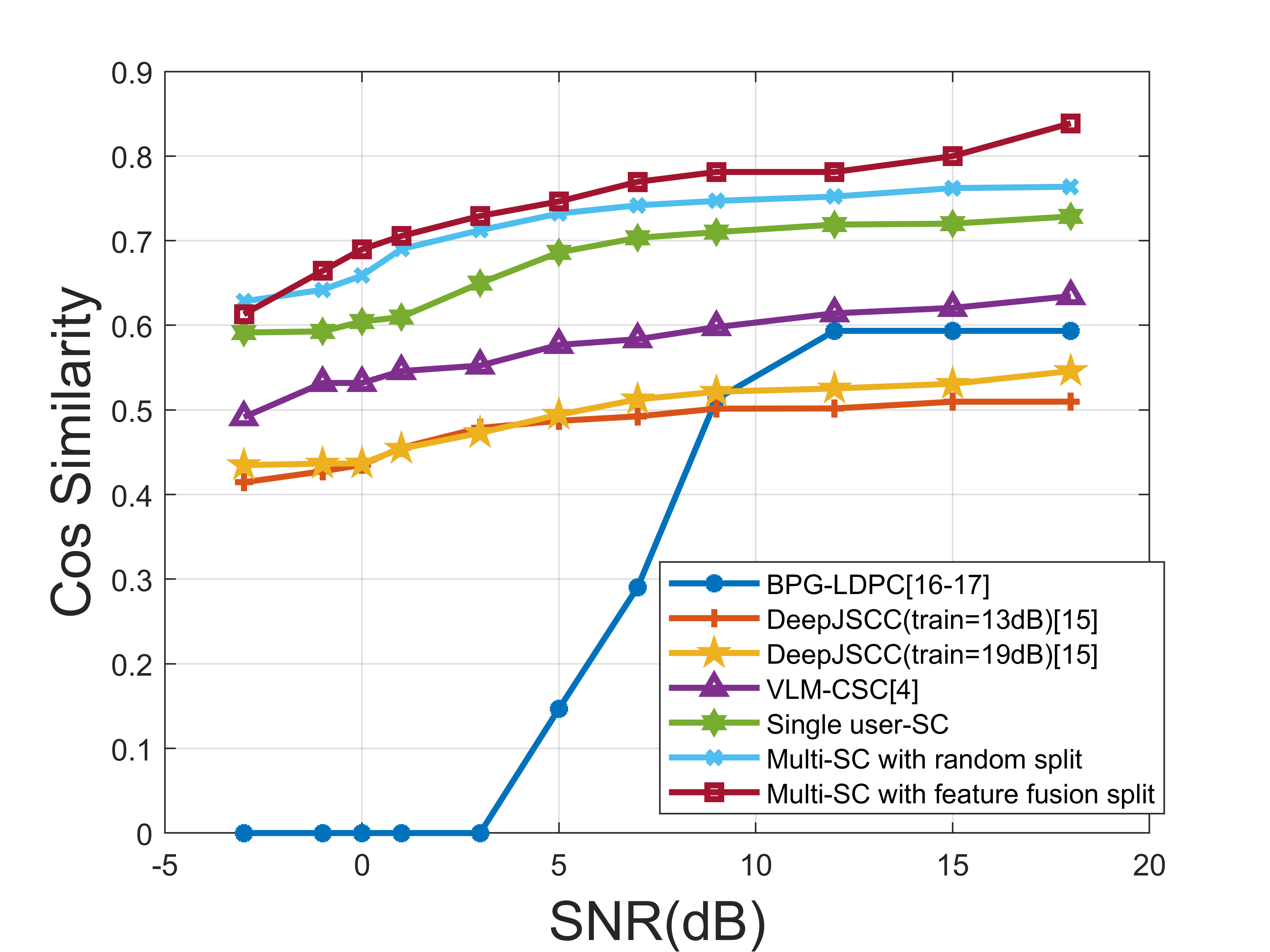}%
		\label{coss_voc_ray}}
	\hfil
	\subfloat[LPIPS versus SNR.]{\includegraphics[width=0.5\linewidth]{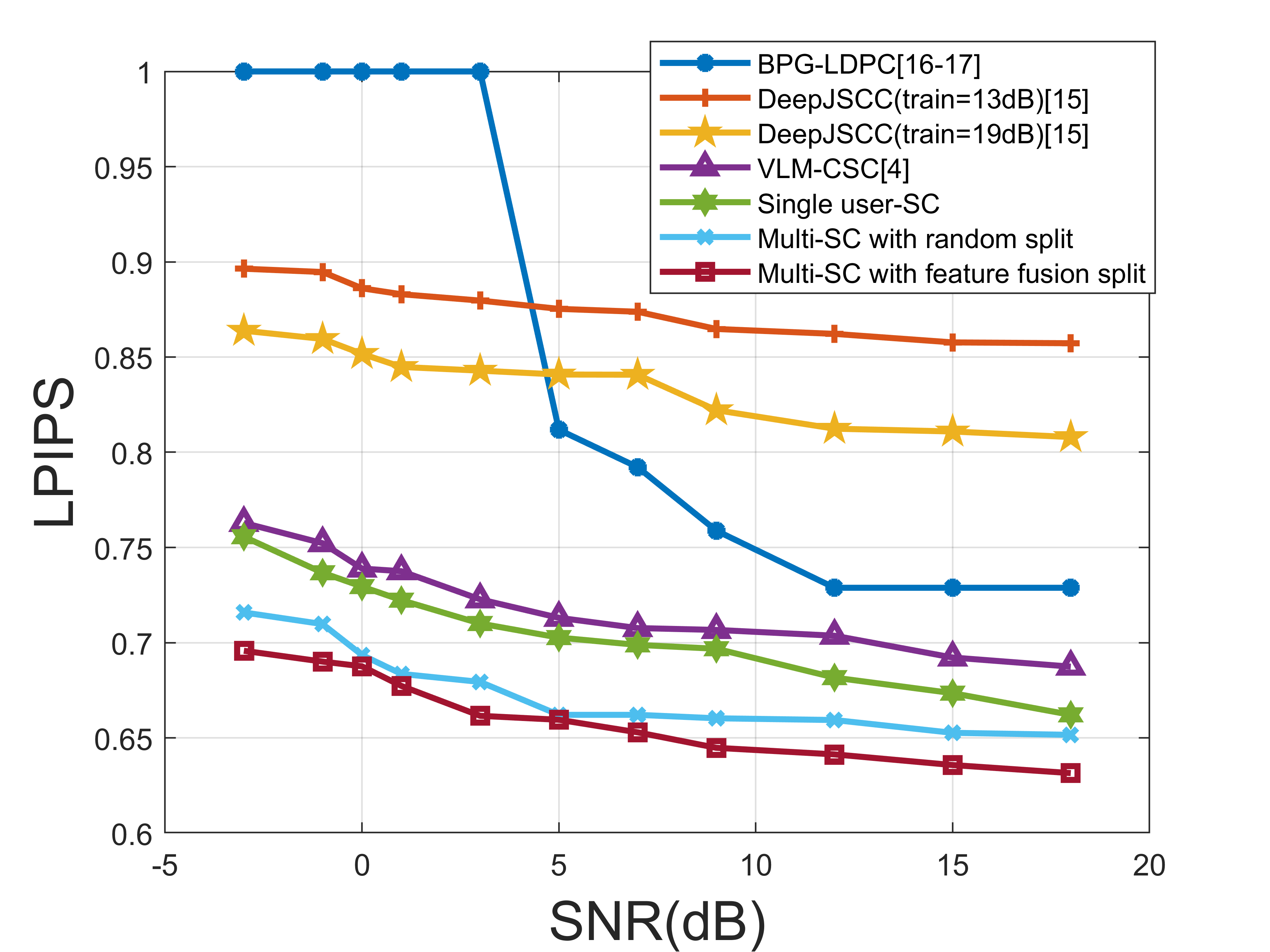}%
		\label{lpips_voc_ray}}
	
	\caption{The results of cosine similarity and LPIPS compared with BPG-LDPC, deep JSCC, VLM-CSC, single text-SC, Multi-SC with random split and baseline Multi-SC with feature fusion split on VOC2012 under Rayleigh.}
	\label{fig_voc_ray}
\end{figure}

\begin{figure}[htpb]\vspace{-0.7cm}
	\centering
	\subfloat[Cosine Similarity versus SNR.]{\includegraphics[width=0.5\linewidth]{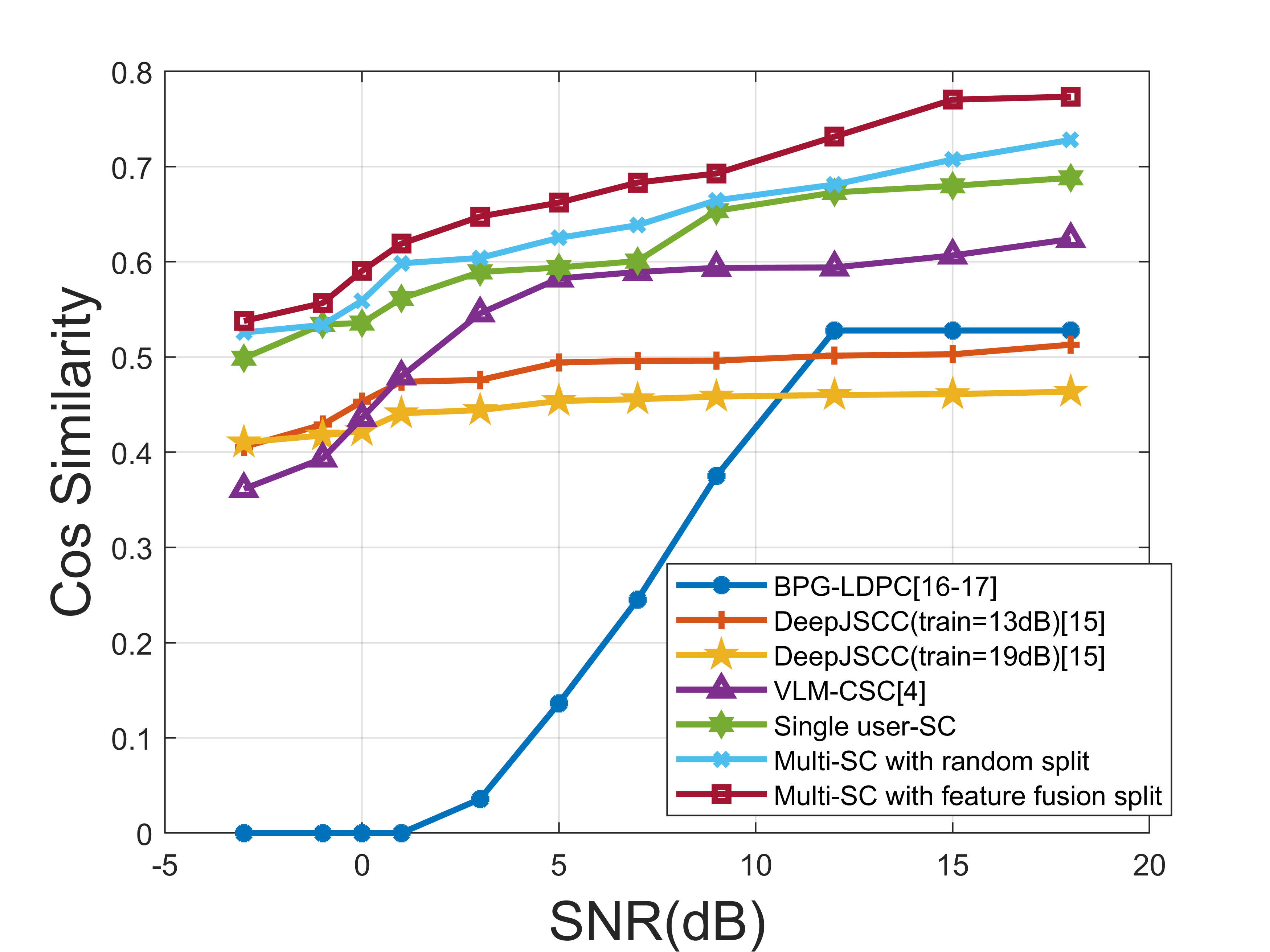}%
		\label{coss_coco_ray}}
	\hfil
	\subfloat[LPIPS versus SNR.]{\includegraphics[width=0.5\linewidth]{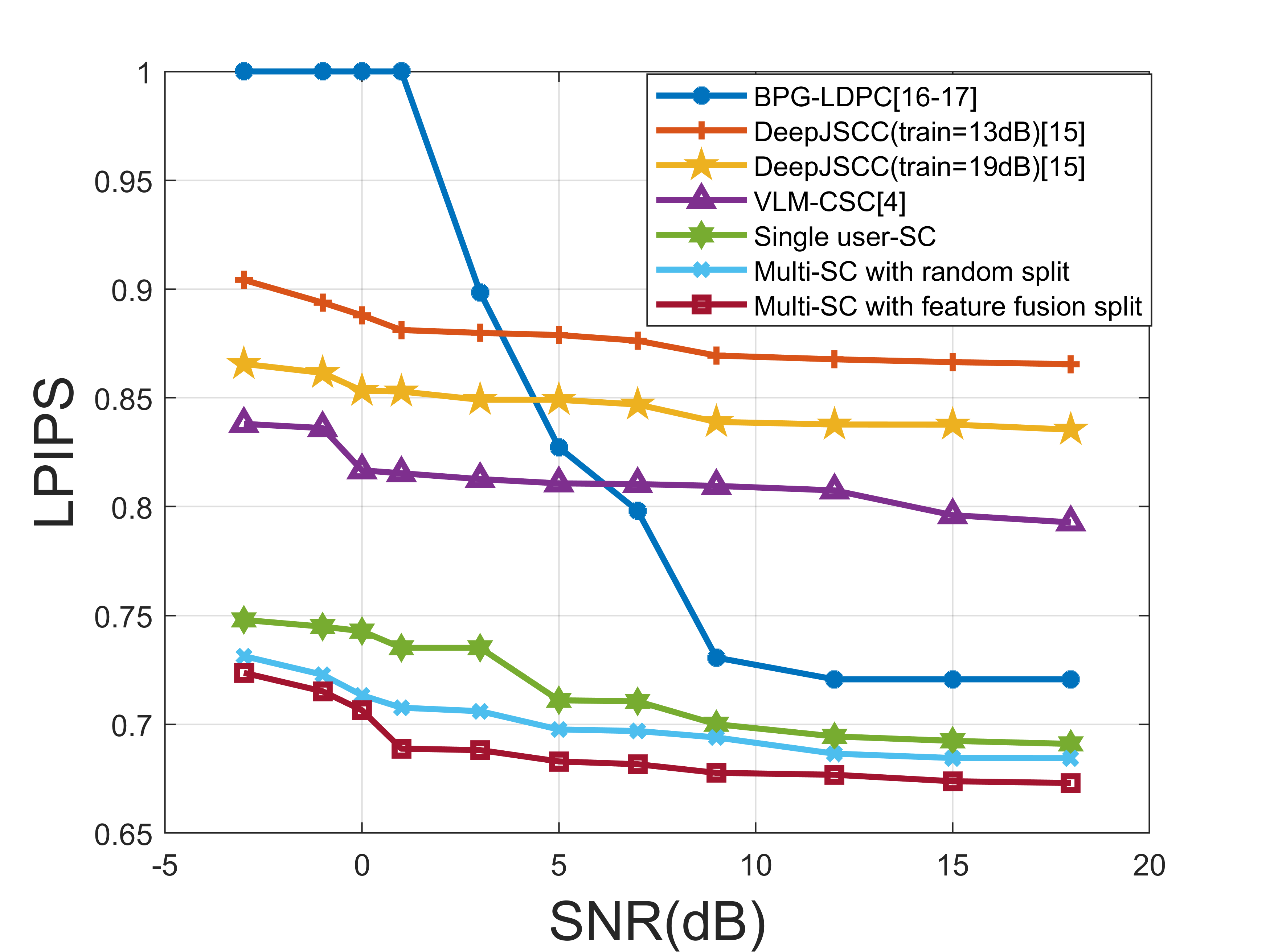}%
		\label{lpips_coco_ray}}
	
	\caption{The results of cosine similarity and LPIPS compared with BPG-LDPC, deep JSCC, VLM-CSC, single text-SC, Multi-SC with random split and baseline Multi-SC with feature fusion split on COCO under Rayleigh.}
	\label{fig_coco_ray}
\end{figure}


\begin{table*}[h]\vspace{-1cm}
	\centering
	\caption{Model Inference time comparison}
	\setlength{\tabcolsep}{2mm}{
		\begin{tabular}{ccccccc}
			\toprule 
                \textbf{Sample source} & \textbf{BPG-LDPC(s)}& \textbf{DeepJSCC(s)}& \textbf{VLM-CSC(s)}& \textbf{Single user SC(s)} & \textbf{\makecell{Multi-SC with \\ random split(s)}} & \textbf{\makecell{Multi-SC with \\ feature fusion split(s)}}  \\
                 \midrule
                Kodak24 & 3 & 2.7  & 17.6 & 6.1 & 7.7 & - \\
                VOC2012 & 3 & 2.7  & 17.6 & 6.1 & 8.4 & 6.7 \\
                COCO & 3 & 2.7  & 17.6 & 6.1 & 6.5 & 5.2 \\
			\bottomrule
	\end{tabular}}
	\label{runing time}%
\end{table*}%

The proposed multi-SC with feature fusion split achieves better performance on the cosine similarity and is robust under AWGN and Rayleigh channels, which indicates that the reconstructed image has a better match with the source image. When there is no correlation between the two images, the cosine similarity score is 0. As shown in Figs. 3(a), 4(a) and 5(a), when the SNR is less than 3 dB, the score of the BPG-LDPC is exactly 0 and the transmitted images cannot be completed. In Fig. 3(a), when the SNR reaches 9 dB, the cos similiarty score reaches 0.7, and the scores of BPG-LDPC and the baseline are similar, indicating that our baseline can effectively transmit the semantic information and reconstruct the source image. In Fig. 4(a), when SNR greater than 9 dB, the score of multi-SC with feature fusion split  is still 4\%-35\% higher than all models, indicating that our model is stronger in different noise channels. Since the test images contain less extracted text details, the VLM-CSC model that relies only on text transmission exhibits significant randomness in the image reconstruction, resulting in a poor performance and the cosine similarity remaining between 0.6 and 0.7, as for Rayleigh signals and AWGN channels.

The lower the LPIPS score, the better the performance. For example, in Fig. 3(b), the LPIPS score of the BPG-LDPC model with less than SNR = 3 dB is very close to 1, indicating that the image quality transmitted by the traditional image transmission model is poor in a noisy environment. As shown in Fig. 5(b), our method achieves significant improvement compared to other methods: under training SNR = 19 dB, an 19\% improvement over deep JSCC, an 8\% improvement over VLM-CSC, a 15\% improvement over BPG-LDPC, a 3\% improvement over single-text SC, and a 2\% improvement over baseline respectively. Although in Fig. 4(b), when the SNR is 12 dB, the LPIPS score of BPG coincides with that of our method. In Fig. 8(b), our method also improves by at least 2\%-22\% on the Rayleigh channel. Although in Fig. 4(b), when the SNR = 12 dB, the LPIPS score of BPG-LDPC is consistent with  our method. Overall, our method still shows a 2\% improvement, demonstrating that the image reconstructed by our method is very similar to the source image. This indicates that our method can achieve a better performance.

However, in terms of inference time, the diffusion model has a higher complexity than the traditional model due to the iterative generation process and the need to gradually refine the noise, resulting in a longer inference time~\cite{11002717,yang2024survey}. Therefore, the inference time is not greatly improved compared with the traditional image transmission model. However, among the three models based on diffusion, the Multi-SC with feature fusion split proposed by us still has certain advantages. On the COCO dataset, our model has a 20\% speed improvement compared to baseline, a 14\% speed improvement compared to single-user SC, a 70\% improvement compared to VLM-CSC, and a similar improvement on the VOC2012 dataset. Since the Kodak24 dataset cannot be partitioned on object detection model, the relevant training data is missing on the Multi-SC with feature fusion split.

\section{Conclusion}
\label{conclusion}
In this paper, we propose a VLM assisted multi-text semantic communication system by exploring multiple textual semantics, which greatly reduces the transmission amount of semantic features without reducing the effective extraction rate of features and realizes image transmission under low SNR conditions. 
We extract the image features by FCN and fuse the adjacent main features into one through the proposed fusion feature method. We extract the text features by VLM, and then use BART to correct spelling errors for reducing the impact of noise on the text. Finally, the image is reconstructed by combining the image features and texts to finish a complete image transmission process.
The simulation results show that feature fusion can dramatically improve the accuracy of semantic features, and accurate text features with a small number of image features can robustly complete the image transmission, and a small number of image features can effectively limit the randomness of the generated images. 

\bibliographystyle{IEEEtran}

\bibliography{reference}

\vfill

\end{document}